**Type of Presentation**:  
Oral: ☐   In-person: ☒  
Poster: ☒   Virtual in Zoom: ☐  
The same: ☐

**Topic: Video processing**

# A New Real-World Video Dataset for the Comparison of Defogging Algorithms

A. Duminil [1], J. P. Tarel [1] and R. Brémond [1]

[1] Université Gustave Eiffel, Laboratoire COSYS-PICS-L, F-77454 Marne-la-Vallée, France
E-mail: alexandra.duminil@univ-eiffel.fr

**Summary:** Video restoration for noise removal, deblurring or super-resolution is attracting more and more attention in the fields of image processing and computer vision. Works on video restoration with data-driven approaches for fog removal are rare however, due to the lack of datasets containing videos in both clear and foggy conditions which are required for deep learning and benchmarking. A new dataset, called REVIDE, was recently proposed for just that purpose. In this paper, we implement the same approach by proposing a new REal-world VIdeo dataset for the comparison of Defogging Algorithms (VIREDA), with various fog densities and ground truths without fog. This small database can serve as a test base for defogging algorithms. A video defogging algorithm is also mentioned (still under development), with the key idea of using temporal redundancy to minimize artefacts and exposure variations between frames. Inspired by the success of Transformers architecture in deep learning for various applications, we select this kind of architecture in a neural network to show the relevance of the proposed dataset.

## 1. Introduction

Visibility restoration of single hazy images is a well-known problem in computer vision and image processing. Visual artifacts in the images such as loss of contrast and color shift, contribute to reduced scene visibility which is detrimental to the performances of computer vision tasks such as segmentation, object/scene detection and recognition. Therefore, dehazing algorithms are often needed as a pre-processing step. Many efficient methods are available for single image dehazing, with or without learning. Several datasets are also available with foggy images and a clear image for reference. Due to the scarcity of foggy scenes, most of these databasets are produced by adding a synthetic fog. And few works tackle video dehazing.

Working on fog removal in videos is more complicated due to lack of foggy video datasets with a clear video for reference. There are single image and video datasets with synthetic fog. For example, Ren et al. [1] generated a video dataset from the NYU-Depth V2 dataset [2], consisting of video sequences of indoor scenes. However, due to the domain gap between synthetic and real world foggy dataset, the results on real world images are not convincing. Recently, Zhang et al. [3] proposed an useful hazy video dataset for learning and data evaluation, called REVIDE. Hazy scenes are produced with a fog machine which is more realistic than synthetic fog.

In order to enrich the state of the art, a new video dataset with and without fog is proposed. It includes video sequences of a reduced-scale but real-world scene with several fog densities and several lighting conditions. Earlier works adressed video visibility restoration as an extension of image restoration. However, applying image processing methods to videos introduces artifacts and flickering effects, due to the lack of temporal consistency. The challenge is to get a video restoration algorithm ensuring spatial and temporal coherence. For instance, Wang et al. [4] and Zhang et al. [3] proposed a video restoration architecture with a deformable Convolutional Neural Network (CNN) module, which achieve interesting results. To deal with the temporal information, recent works have been inspired by the Transfomers architecture [5] which gives interesting results in video restoration for super resolution and image synthesis applications. While many methods use pure Transformers to solve their tasks, we propose to combine CNN and Tranformer architectures to process the temporal information in videos. We have investigated including a Transformer module in a neural networks architecture within a video restoration method. The proposed algorithm, called TCVD, is demonstrated with the proposed VIREDA dataset.

## 2. Foggy video dataset

### 2.1. Creation process

The main idea was to collect videos with and without fog as well as associated depth maps. This was achieved by creating a fog chamber to reproduce a hazy scene with the help of a fog machine. This fully controllable system allows to roughly control the fog density and illumination conditions. A series of static objects were included in the chamber, together with a small scale car which was remotely operated to introduce temporal variations in the otherwise static scene. In order to make sure that the position of the car was exactly the same for all visibility and lighting conditions, a stop-motion technique was implemented. The following acquisition process was applied for each robot position: First, a video of the scene without fog was recorded while alterning the different lightings. Next, fog was fed into the chamber. Then, after a few seconds of waiting for the fog to settle, the video recording was started at the same time as the venting system was started. As fog was slowly evacuate, series of frames with decreasing densities of fog were recorded with each lighting. At each position of the robot, videos with different densities of fog, including without fog, have been acquired with a Kinect v2. This device captures both images and depth maps of the scene thanks to a Time of Flight technology, which provides additional information that are useful for some fog removal algorithms. At the end of the process, the frames corresponding to each illumination condition and fog density were splitted, and recomposed into new videos were the car was moving in stable visibility and lighting conditions.

### 2.2. Dataset content

The use of LED strips made it possible to produce different illumination conditions. To simulate heterogeneous lighting (such as at night), lamps with different color temperatures were used. In the other five lighting conditions, white LEDs were used to simulate daylight. Fog density was evaluated by monitoring the contrast in a black & white panel. Three fog densities are available in the dataset (see **Fig. 1**), with contrasts of 0.015, 0.05 and 0.15 (the lower the contrast, the higher the density).

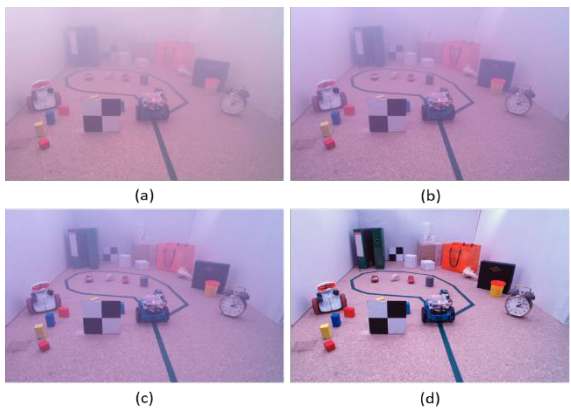

**Fig. 1.** Frames with three different density : a) foggy frame with c = 0.015, b) foggy frame with c = 0.05, c) foggy frame with c = 0.15, d) ground truth.

The dataset contains 18 foggy videos and 6 ground truth videos, each containing an average of 180 frames (i.e. about 7 seconds). Additionally, there are depth maps associated with each position of the robot. The dataset is available on: https://github.com/alex-dml/VIREDA-video-dehazing

### 2.3. Depth maps

The use of the Kinect made it possible to collect depth maps via a Time of Flight (ToF) sensor. This sensor measures the time the light takes to travel a distance between the signal emission and its return to the sensor. As it is not sensitive to change in brightness, one depth map per robot position was enough. Depth maps are an additional source of information in the database. **Fig. 2** shows the depth map for a robot position and the hazy image.

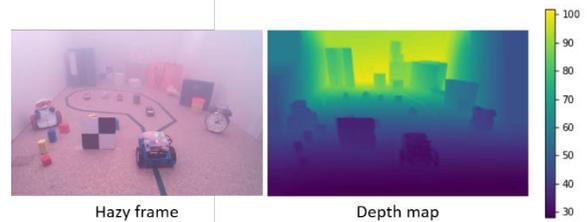

**Fig. 2.** Hazy frame and the associated false-colour depth map. Graduations are in centimeters.

## 3. Video defogging method

The popularity of the transformer architecture prompts exploration of its use for different types of tasks. Few works are yet inspired by this model for fog removal applications in videos. In this section, we present a multi-step method, called TCVD (Transformer-CNN architecture for Video Defogging), to process spatio-temporal information in video sequences. At each stage, spatial and temporal information are processed through several blocks and are then merged. This method combines two different architectures: CNNs to process spatial information and Transformers to ensure temporal consistency.

The overall architecture is based on an auto-encoder inspired by U-net. Using the transformer architecture is suitable to deal with sequences of data due to its ability to bind data within a sequence and to parallelize tasks. However, the Transformer has some drawbacks including a low inductive bias. The inductive bias represents the set of assumptions made by the model that allows it to learn and generalize beyond the learning data. On the other hand, CNNs have a strong inductive bias, allowing the model to achieve interesting performances with less training data. However, CNNs do not allow capturing the interdependence of images within a video sequence. The joint use of these two architectures can be an advantage.

### 3.1. Encoder

The encoder part of our architecture consists of four CNN modules and three modules based on a Transformer architecture, called TpFormer. **Fig. 3** represents the overall diagram of the encoder.

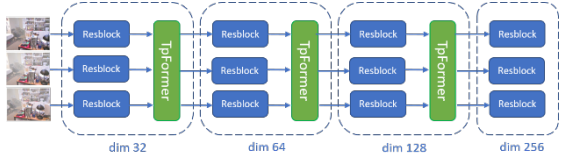

**Fig. 3.** Encoder architecture of the proposed method with three successive frames as input.

Each step corresponds to the feature extraction of each image of the sequence. The features are then merged and form the input of the TpFormer blocks at several dimensions (which correspond to 2D convolution layer filters: 32, 64, 128, and 256). At the end of the feature extraction steps, the successive image triplets in the video are concatenated along the time axis.

### 3.2. The TpFormer module

The TpFormer module, detailed in **Fig.4**, is composed of several layers including a Multi-Head-Attention (MHA) layer [5]. This layer makes it possible to carry out different attention calculations in parallel. Using attention mechanism allows the network to learn the sequential dependency between frames. The features of the resulting frames are then dissociated and concatenated to the resulting spatial features of the CNN block of the same step. The merged information is the input to the next step.

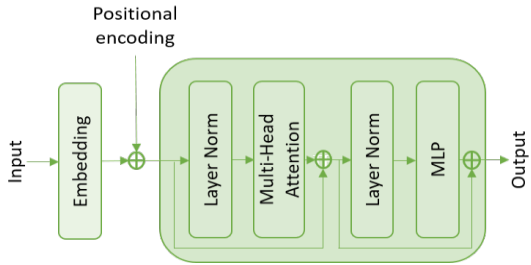

**Fig. 4.** TpFormer module, inspired by ViT [5].

### 3.3. Decoder

Once the encoder has extracted all the spatio-temporal characteristics of the images of the sequence, the decoder part will, symmetrically to the encoder, over-sample the encoded data of the central image, so as to obtain an image of dimensions equal to that of the input images. To upsample, Conv2DTranspose layers (Keras) were used, as well as skip connections from the encoder to retain fine details in the images.

### 4. Implementation

The implementation of the algorithm was carried out with the Keras/Tensorflow frameworks. We use the ADAM optimizer and a learning rate of 0.0001.

The Loss function is a combination between the L1 and SSIM distances. It is defined as follows:

$$L = a * L_{ssim} + b * L1, \quad (1)$$

where $a$ and $b$ are balanced coefficients. We have resized foggy inputs and ground truths to a size of 224 x 224 and augmented them with horizontal flip and random 90° rotations.

### 5. Evaluation

Very few methods of video defogging provide publicly available code. Accordingly, our video defogging algorithm, still under development, only served to show some preliminary results on our dataset. TCVD is trained on the REVIDE dataset and TCVD2 is trained on both the REVIDE dataset and the synthetic images dataset (inspired by [1]) in order to have various types of data and be able to generalize. Then, we tested that with our video dataset, VIREDA.

### 5.1. Quantitative evaluation

The proposed algorithm is compared to different state-of-the-art algorithms, including two prior-based methods (DCP [7] and MFP [8]) and a learning-based methods (FFA-Net [9]). Due to lacks of available codes for video dehazing, only image dehazing methods are used for comparison. Therefore, only the visual quality of the restored frames is assessed in this evaluation. The FFA-Net algorithm is retrained on the REVIDE dataset. **Table 1.** shows sample of the quantitative results with the SSIM [6] and PSNR metrics in the case of three fog densities including the different lighting conditions. All frames are resized to 224 x 224. The results illustrate that MFP, FFA-Net, TCVD and TCVD2 methods are competitive. The results obtained with the DCP method can be explained with the visual colorshift results seen in the **Fig. 5**.

| Fog density | DCP | MFP | FFA | TCVD | TCVD2 |
|---|---|---|---|---|---|
| 0.015 | 0.57 | **0.74** | 0.68 | **0.72** | 0.70 |
|  | 9.89 | **15.92** | 14.14 | 13.29 | **15.95** |
| 0.05 | 0.60 | **0.80** | 0.71 | **0.76** | 0.74 |
|  | 10.30 | **16.09** | 14.35 | 13.97 | **15.73** |
| 0.15 | 0.66 | **0.86** | 0.78 | 0.79 | **0.82** |
|  | 11.35 | **16.56** | 16.22 | 14.68 | **17.16** |

**Table 1.** Quantitative evaluation on different fog densities with the SSIM (top) and PSNR (bottom) metrics.

MFP method provides better results than those of our TCVD method with two fog densities : 0.05 et 0.15. Despite this, **Fig. 5** shows that MFP is not as efficient as TCVD to removes fog. However, TCVD introduces black artifacts which penalize the results. TCVD2 shows better results with PSNR metric than MFP. Training TCVD2 on both databasets appears to reduce artifacts introduced when training with REVIDE only.

### 5.2. Qualitative evaluation

**Fig. 5** shows sample images of the dataset restored with different dehazing methods. Each row in the figure corresponds to a different illumination condition with a fog density of 0,05. The DCP method seems to deteriorate the visual quality of the images, as shown by the different color shifts. The MFP and FFA-Net methods allow to slowly attenuate dense fog but produce more satisfactory results with lighter fog. The TCVD method, intended for videos, attenuates fog well but leaves black artifacts. This effect may be caused by the high intensity point lights implemented to simulate daylight: the algorithm may interpret these areas as fog to remove. As we noticed in the previous section, TCVD2 produces better quality results than TCVD. The artificial lighting used to create the video dataset produced slightly teinted images despite the use of white LEDs. It produced a colored fog which seems to disturb the restoration process by introducing some artifacts and color shifts.

### 6. Conclusion

A new publicly available test dataset has been generated for the testing of video defogging. Although it is only composed of a single scene, it is available with several lighting conditions and several fog densities, and provides depth maps which may help assessing the quality of some defogging algorithms. For deep learning algorithms, it is useful to propose new sets with different conditions to expect a good generalization. We have implemented a multi-step hybrid method, called TCVD, which combines two architectures: CNNs and Transformers. This method helps reducing fog in realistic videos. However, the small number of publicly available code on defogging methods for videos does not allow to realize a complete benchmark.

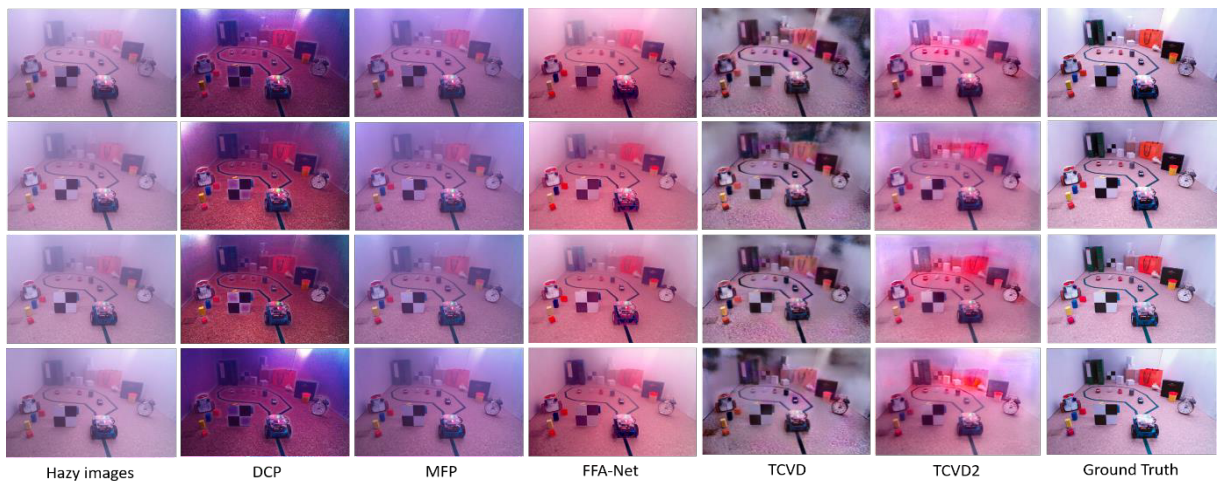

**Fig. 5.** Qualitative evaluation of dehazing algorithms on the VIREDA dataset with 0.05 as fog density value.